\documentclass{article}

\usepackage{amsmath}
\usepackage{appendix}
\usepackage{graphicx}
\usepackage{hyperref}
\usepackage{longtable}
\usepackage{multirow}
\usepackage{subcaption}
\usepackage{tabularray}
\usepackage{url}
\usepackage{xcolor}
\usepackage{natbib}

\makeatletter
\renewcommand\@date{}
\makeatother

\usepackage[a4paper, margin=1.2in]{geometry}

\usepackage{fancyhdr}
\title{Multispectral Fine-Grained Classification of Blackgrass in Wheat and Barley Crops}
\author{Madeleine Darbyshire$^{1}$ \and Shaun Coutts$^{2}$ \and Eleanor Hammond$^{3}$ 
        \and Fazilet Gokbudak$^{4}$ \and Cengiz Oztireli$^{4}$ \and Petra Bosilj$^{1}$
        \and Junfeng Gao$^{2}$ \and Elizabeth Sklar$^{2}$ \and Simon Parsons$^{1}$}

\begin{document}
\maketitle

\footnotetext[1]{MD, PB and SP are with the Lincoln Centre for Autonomous Systems, University of Lincoln, UK.
        {\tt $\{$mdarbyshire, pbosilj, sparsons$\}$@lincoln.ac.uk}}%
\footnotetext[2]{SC, JG and ES are with the Lincoln Institute of Agri-food Technology, University of Lincoln, UK.
        {\tt $\{$scoutts, jugao, esklar$\}$@lincoln.ac.uk}}%
\footnotetext[3]{EH is with the Department of Biology, University of Oxford, UK.
        {\tt eleanor.hammond@biology.ox.ac.uk}}%
\footnotetext[4]{FG and CO are with University of Cambridge, UK.
        {\tt $\{$fg405, aco41$\}$@cam.ac.uk}}%

\begin{abstract}
As the burden of herbicide resistance grows and the environmental repercussions of excessive herbicide use become clear, new ways of managing weed populations are needed. This is particularly true for cereal crops, like wheat and barley, that are staple food crops and occupy a globally significant portion of agricultural land. Even small improvements in weed management practices across these major food crops worldwide would yield considerable benefits for both the environment and global food security. Blackgrass is a major grass weed which causes particular problems in cereal crops in north-west Europe, a major cereal production area, because it has high levels of herbicide resistance. Moreover, detecting blackgrass in grass crops is challenging due to its visual similarity to these crops. Despite this, a systematic review of the literature on weed recognition in wheat and barley, included in this study, highlights that blackgrass – and grass weeds more broadly – have received less research attention compared to certain broadleaf weeds. With the use of machine vision and multispectral imaging, we investigate the effectiveness of state-of-the-art methods to identify blackgrass in wheat and barley crops. As part of this work, we present the Eastern England Blackgrass Dataset, a large dataset with which we evaluate several key aspects of blackgrass weed recognition. Firstly, we determine the performance of different CNN and transformer-based architectures on images from unseen fields. Secondly, we demonstrate the role that different spectral bands have on the performance of weed classification. Lastly, we evaluate the role of dataset size in classification performance for each of the models trialled. All models tested achieved an accuracy greater than 80\%. Our best model achieved 89.6\% and that only half the training data was required to achieve this performance. Our dataset is available at: \url{https://lcas.lincoln.ac.uk/wp/research/data-sets-software/eastern-england-blackgrass-dataset/}.
\end{abstract}

\section{Introduction}
Weeds compete with crops for light, water and nutrients. Additionally, grains from grass weeds get harvested alongside grains from the crop, contaminating the harvest, and can drastically lower the price received for the crop. Current mainstream weed management practices involve broadcast spraying, where the entire field is sprayed with a selective herbicide. An alternative approach, known as precision spraying, targets herbicide only at areas of weed cover guided by machine vision. Precision spraying facilitates up to 95\% reduction in herbicide use (depending on crop and field) \cite{zanin2022reduction}, and a proportional decrease in herbicide costs and environmental damage from runoff. In addition, precision spraying allows a wider range of herbicides (potentially broad spectrum) to be used, opening more options for resistance management \cite{busi2020rotations}. Meanwhile, other technologies, like automated camera-guided mechanical weeding, promise non-chemical weed control. Both of these promising technologies are predicated on accurate, real-time weed recognition within often complex crop canopies.

Efforts to reduce herbicide usage in staple cereal crops have the potential to deliver significant impact. The area of global cropland devoted to growing cereals, and in particular rice, wheat and maize, is orders of magnitude greater than that of many vegetable crops \cite{fao}. Furthermore, wheat is second only to rice as a global staple \cite{Eren2022}, with a global consumption of 65.6 kg per person per year. Therefore, not only is sustainable and reliable crop care for wheat critical for global food security, any technology that reduces herbicide usage in these global staples has the potential to massively reduce the environmental impact of herbicide use globally \cite{darbyshire2024review}.

Moreover, particular weeds present a greater threat to global food supply than others; grass weeds are a particular problem in wheat production due to their biological similarities \cite{chhokar2012weed, young1996weed, moyer1994weed, Hick2018}. In Europe, blackgrass is one of the most economically damaging weeds \cite{Vara2020}, so effective strategies to manage populations are a priority. Due to high levels of resistance to selective herbicides \cite{Hick2018}, using them for precision spraying may not be sufficient to control blackgrass populations. In these instances, precision spraying with non-selective herbicides could be an alternative approach, but requires a high degree of confidence in identifying and targeting so as not to damage the crop. Such confidence will rely on well tested, accurate, real time machine vision systems that can be deployed at scale to the field.

\subsection{Paper Overview}
To facilitate evaluating agricultural vision systems on in-context, in-field data, we present the Eastern England Blackgrass Dataset, a weed image data set featuring wheat and barley of over 15,000 images across 51 fields in the UK, with and without blackgrass --- a grass weed of cereal crops.

Unlike other crop and weed datasets, this dataset focuses on recognising the presence of blackgrass in wheat. The visual similarity of blackgrass to cereal crops poses a challenging fine-grained classification task for precision weed management technology. The release of this dataset will accelerate the development of weed identification systems for blackgrass, and due to their visual similarity, other grass weeds, for wheat, a critical global staple.

The contributions of this paper are as follows:
\begin{itemize}
  \item A dataset containing over 15,000 multispectral images of wheat and blackgrass stratified by field, season, geolocation and soiltype, enabling the study of domain generalisation with respect to these factors;
  \item An evaluation the effect of using different spectral bands for blackgrass classification; and
  \item An assessment of the effect of training data quantity on model performance providing guidelines for future dataset building.
\end{itemize}

\section{Background}
We start by reviewing the relevant research. First we look at how the variety of color spaces have been used in the research to date, and secondly, how those color spaces, along with other features, have been used to recognise weeds. Lastly, we systematically evaluate all recent research on weed recognition in wheat and barley to understand on which weed species research effort has been focused.

\subsection{Color Spaces for Weed Recognition}
 Initially, color indices derived from RGB like Excess Greenness (ExG), Excess Redness (ExR), ExG minus ExR (ExGR) and Normalized Difference Index (NDI) were used in weed recognition research \cite{jensen2021predictive}. Multispectral imaging for crop and weed discrimination is now an active area of research. There is an existing precedent for the use of multispectral imaging in plant analysis since the Normalized Difference Vegetation Index (NDVI) \cite{rouse1974monitoring}, derived from near-infrafed (NIR) and red light, has long been used in agriculture to assess crop health. Studies trialling different combinations of NIR, Red and NDVI as inputs to a convolutional neural network (CNN) to detect weeds in sugar beet fields found that NIR and Red values used together performed the best when identifying weeds \cite{sa2018weednet, razaak2019integrated}. Interestingly, including NDVI as an input, alongside NIR and Red values, did not improve the performance \cite{sa2018weednet, razaak2019integrated}. With the use of hyperspectral cameras, it was found that increasing the number of bands from 3 up to 61 resulted in incremental improvements in weed species classification accuracy \cite{farooq2018analysis}. In a study where several color indices where trialled, the inclusion of NIR information either alone, or alongside other indices like ExG and NDI, significantly improved the segmentation accuracy \cite{wang2020semantic}.

\subsection{Weed Recognition Techniques}
Early research into techniques for weed recognition relied on hand-crafted feature-based techniques. Approaches to discriminating grass weeds in cereal crops use color \cite{anderegg2023farm}, texture \cite{xu2020recognition}, depth \cite{xu2020recognition} and spectral \cite{amziane2021weed, jensen2021predictive} features. For weed classification, support vector machine (SVM) \cite{cristianini2000introduction} and k-Nearest Neighbor (k-NN) \cite{cover1967nearest} approaches were used to learn decision boundaries in the feature space \cite{xu2020recognition}. 

Deep learning-based weed classification rose to prominence after the success of early CNN architectures like AlexNet \cite{krizhevsky2012imagenet} and VGG \cite{simonyan2014very}. However, ResNet \cite{he2016deep} improved the applicability of this technology by utilising skip connections to address the vanishing gradient problem, which enabled the training of deeper neural networks, resulting in improved performance and convergence. In the following years, deep learning-based methods with faster inference speeds and smaller memory footprints enabled weed recognition on the edge, like the popular MobileNetV3 architecture, which exploits lightweight depthwise separable convolutions for improved efficiency \cite{howard2019searching}. EfficientNet employed a neural architecture search approach to design a network and a novel method to scale the model depth, width, and resolution of the network in order to balance accuracy and efficiency\cite{tan2019efficientnet}. Recently, transformers have demonstrated how powerful self-attention mechanisms can be for recognizing complex patterns in sequential data \cite{vaswani2017attention}. Vision Transformer (ViT) \cite{dosovitskiy2020image} applied transformer technology to image classification by treating an image as a sequence of patches, which were then linearly embedded and processed through a transformer architecture. The self-attention mechanism in transformers allowed ViT to analyze global contextual information, improving performance on various image classification benchmarks. 

ResNet, MobileNetV3 and MobileViT \cite{mehta2021mobilevit} --- a compact version of ViT --- demonstrated strong performance in weed classification in wheat \cite{alirezazadeh2023comparative}. Deep learning-based object detection techniques, that identify the region where an object is located in an image, have been used to detect grass weeds in wheat crops including using DetectNet \cite{dyrmann2017roboweedsupport}, SSD \cite{dyrmann2018using}, Faster R-CNN \cite{xu2021multi, xu2023weedsnet}. Additionally, there is work segmenting grass species, such as ryegrass, using an adapted ERFNet \cite{su2021real}, as well as green foxtail and horsetail using an adapted U-Net network \cite{zou2021modified}.

\subsection{Weed Recognition Datasets}
In recent years, there have been several popular crop and weed datasets. Some have image-level labels such as the Plant Seedlings Dataset \cite{giselsson2017public}, some have bounding box labels like the Lincoln Beet dataset \cite{salazar2022beyond} and some have pixel-level labels like the CWFID dataset \cite{haug2015crop}. As the granularity of the labels has increased, from image-level to pixel-level, it has increased the precision with which that weed can be identified and thus treated. In this landscape, we see the problem of identifying grass weeds in cereal crops is generally overlooked. This is despite the identificiation of grass weeds in cereal crops often being more challenging than other weed detection problems, due to significant occlusion, even during early growth stages, and the visual similarity between the crop and weed.

\subsection{Prevalence of Grass Weed Recognition in Cereal Crops}
We systematically reviewed the latest machine vision research on wheat and barley crops from 2015 and identified that grass weeds are understudied compared to broadleaf weeds. Grass weeds are more challenging to distinguish from cereal crops (which are themselves grass species). Papers were identified in Google Scholar and Scopus using search query ``weed identification'' OR ``weed recognition'' OR ``weed discrimination'' OR ``weed detection'' OR ``weed classification'' OR ``weed segmentation'' on 20th October 2023. All publications from recognised, peer-reviewed, English-language journals and conference venues that could be accessed were included. For this analysis, we are only interested in weed recognition in wheat and barley crops, so excluding all other weed recognition papers we found a total of 79 papers. 

\begin{figure*}[htbp]
  \centering
  \includegraphics[width=\linewidth]{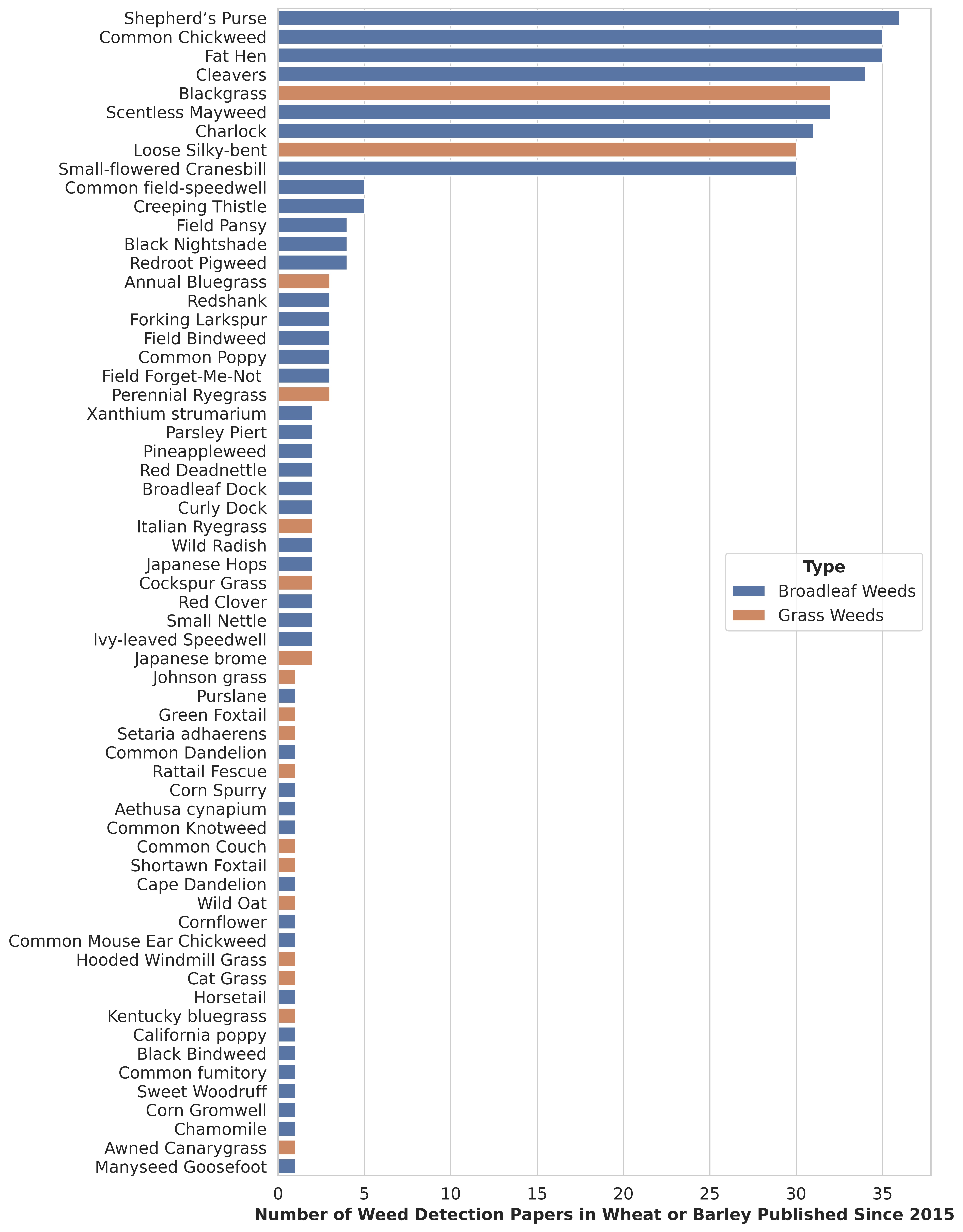}
  \caption{Weed recognition research effort in wheat and barley across weed Species, as measured by published papers, for broad leaf (blue) and grass (red) weeds.}
  \label{fig:papers-weeds}
\end{figure*}

Our analysis of the distribution of weeds species used in weed recognition papers in wheat and barley crops is shown in Figure \ref{fig:papers-weeds}. Each weed species from each paper is added to the cumulative total for that species. The weed species are divided into eudicots (broadleaf plants) and monocots (grasses). While the common names are used in the figure, the full list of papers categorised by weeds, along with their Latin names, is available in Appendix \ref{app:weed}. 

Figure \ref{fig:papers-weeds} shows that broadleaf weeds have garnered more research interest than grass weeds. One possible explanation for this discrepancy is how challenging weed recognition is in grass weeds compared to broadleaf weeds. Grass crops also present a high level of challenge compared to broad leaf crops as they occlude one another, as well as any weeds in the field, even at relatively early growth stages. By contrast, many broadleaf crops remain spaced out, display distinctive features, for example distinctive color and leaf shape, at early growth stages. Additionally, many grass crops look very similar to other grass plants. Distinguishing between grass species is an example of \emph{fine-grained visual classification}. This is a subgroup of classification tasks that aims to distinguish between similar examples based on some subcategory like plant species or car model \cite{ma2021research}. Moreover, like with the classification of medical images, another fine-grained visual classification task, experts are required to produce ground-truth images --- adding to the cost of producing a dataset. Therefore, grass weed recognition in cereal crops has some of the same challenges as medical imaging domains when it comes to building datasets.

\begin{figure*}[t]
\centering
\subcaptionbox{Wheat}
{\includegraphics[height=1.2in]{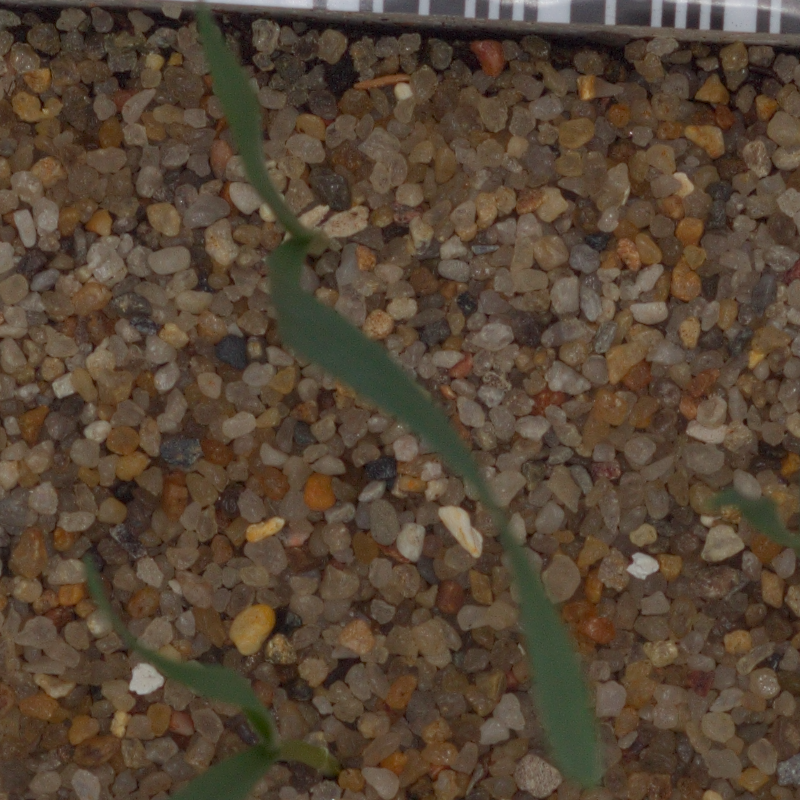}}
\subcaptionbox{Blackgrass}
{\includegraphics[height=1.2in]{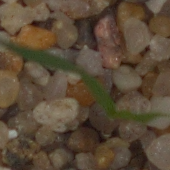}}
\caption{Example images from the Plant Seedlings Dataset \cite{giselsson2017public}.}
\label{fig:psd}
\end{figure*}

An early contribution to the field of deep learning for weed recognition in wheat was the Plant Seedling Classification dataset \cite{giselsson2017public}, shown in Figure \ref{fig:psd}. This dataset contained images of crops including wheat, maize and sugar beet along with 9 common weed species including blackgrass. Over a third of the papers on wheat/weed recognition reviewed for this survey carried out experiments on the Plant Seedling Classification dataset \cite{chavan2018agroavnet, binguitcha2019crops, elnemr2019convolutional, jiang2019novel, espejo2020improving, fu2020field, rahman2020performance, skacev2020development, trong2020late, espejo2021combining, espejo2021testing, farkhani2021weed, mounashree2021automated, siddiqui2021neural, trong2021yielding, damilare2022weed, kumar2022weed, kundur2022deep, mu2022faster, reddy2022weed, srivastava2022role, tara2022machine, bharathi2023plant, dheeraj2023using, guo2023identification, mishra2023comparative, mu2023densenet}. This dataset has made a great contribution to the field by drawing attention to the problem of crop and weed discrimination for crops that are critical to the world's food supply. However, using this dataset does not enable an evaluation of in-field crop and weed discrimination.

Apart from the papers using the Plant Seedling Dataset listed above, the rest of the research papers we reviewed (listed in Appendix \ref{app:weed}) used in-field data. A few papers explore the problem of weed detection in mature wheat crops using both color features \cite{rasmussen2019pre} and LiDAR \cite{shahbazi2021assessing}. Most of the others attempt to detect weeds at an early stage. Many of these focused solely on broadleaf weeds \cite{pflanz2018weed, alirezazadeh2023comparative, el2022metaheuristic, zhuang2022evaluation, almalky2022efficient, almalky2023deep, almalky2023real, xu2023weedsnet} which are easier to distinguish in grass crops than grass weeds. Some of the work discriminating grasses, hand-crafted approaches using color \cite{anderegg2023farm}, texture \cite{xu2020recognition}, depth \cite{xu2020recognition} and spectral \cite{amziane2021weed, jensen2021predictive} features were implemented. In this work, we focus on using deep learning-based approaches as they have been shown to be more robust to changes in illumination and occlusion.



\section{Materials and Methods}
In this section, we first present our dataset of blackgrass in wheat and barley crops. We then outline our classification approach including architectures, experiments, training settings and evaluation methods.

\subsection{Dataset}
Our dataset\footnote{The Eastern England Blackgrass Dataset is available here: \url{https://lcas.lincoln.ac.uk/wp/research/data-sets-software/eastern-england-blackgrass-dataset/}}, the Eastern England Blackgrass Dataset, consists of images of blackgrass in wheat and barley crops collected from 51 fields across eight different soil types in the east of England, a key grain growing region of the UK, between 3rd November 2020 and 16th October 2021.

Fields were found by identifying and contacting land managers across eastern England. We started with those we have worked with previously, and identified new land mangers by introduction from those land managers we were already involved. Land managers were asked if we could access fields that were growing wheat or barley, and contained areas that had both low and high black grass densities. We worked with land managers to find fields across different soil types.

An average of 300 images was taken from each field. The majority of fields had more than 100 images both with and without blackgrass, however, seven fields only had `blackgrass' images and four fields only had `no blackgrass' images.

Images were captured using Micasense RedEdge-MX. The camera was mounted on an image collection rig kept at 95cm throughout data collection. In addition to RGB, the camera captures two additional bands: Red Edge (717 nm) and near-infrared (NIR) (842 nm). Red Edge and NIR wavelengths were included because they are known to aid in plant species differentiation, as variations in leaf structure influence how these wavelengths are reflected \cite{lang2015near}.

As shown in Table \ref{tab:dataset-fields}, the images were split into training, test and validation sets by field so that there is approximately a 80-10-10 split for training, validation and test sets. The aim of this stratification is to ensure that each model's ability to generalise to unseen fields is assessed. Table \ref{tab:dataset-images} shows how many images are in each set according class, crop type and season. Care was taken to ensure the sets were balanced in terms of positive and negative examples as well as to ensure each set contain some images from each stage in the season and each crop type.

\begin{table}[t]
    \centering
    \begin{tabular}{ |c|c| } 
        \hline
        Dataset & Number of Fields \\
        \hline
        Train      & 37 \\ \hline
        Validation & 8  \\ \hline
        Test       & 6  \\ \hline
    \end{tabular}
    \caption{Training, validation and testing is performed on separate sets of images from different fields.}
    \label{tab:dataset-fields}
\end{table}

\begin{table}[t]
    \centering
    \begin{tblr}{ |c|c|c|c|c|c| } 
        \hline
        &  & Train & Validation & Test & Total \\ 
        \hline
        \SetCell[r=2]{c}{Class}     & Blackgrass    & 5770 & 1023 & 1095 & 7888 \\ \cline[dashed]{2-6}
                                   & No Blackgrass & 5982 & 1036 & 1023 & 8041 \\ \hline
        \SetCell[r=3]{c}{Crop Type} & Winter Wheat  & 9425 & 1731 & 1569 & 12725 \\ \cline[dashed]{2-6}
                                    & Triticale  & 172 & 0 & 0 & 172 \\ \cline[dashed]{2-6}
                                   & Spring Barley & 542  & 123  &  372 & 1037  \\ \cline[dashed]{2-6}
                                   & Winter Barley & 1613 & 205  & 177  & 1995  \\ \hline
        \SetCell[r=3]{c}{Season}   & Early         & 3631 & 335  & 583  & 4549  \\ \cline[dashed]{2-6}
                                   & Mid           & 4571 &  841 & 1166 & 6578  \\ \cline[dashed]{2-6}
                                   & Late          & 3550 &  883 & 369  & 4802  \\ \hline
    \end{tblr}
    \caption{Stratification of image numbers for each class, crop and season.}
    \label{tab:dataset-images}
\end{table}

Land managers were consulted to find patches of blackgrass in wheat and barley fields. Images of class `no blackgrass' were taken near the blackgrass patches. Fields with a good number of blackgrass patches were re-visited to collect images at different growth stages of the crop and blackgrass. 
The assessment of ``Blackgrass'' or ``No Blackgrass'' was made at the time of image collection by the same observer for all images across all fields to maximise the number of images collected. 

The observer often needed to carefully examine suspected plants for distinguishing features, such as the absence of auricles and jagged ligules. Since these details are not clearly visible in the images, even an expert observer would struggle to re-identify blackgrass from the photos alone. As a result, the dataset contains only image-level labels.

\begin{figure*}[t]
\centering
\subcaptionbox{Blackgrass: Early Season \label{fig:dataset:bg:early}}
{\includegraphics[width=0.4\linewidth]{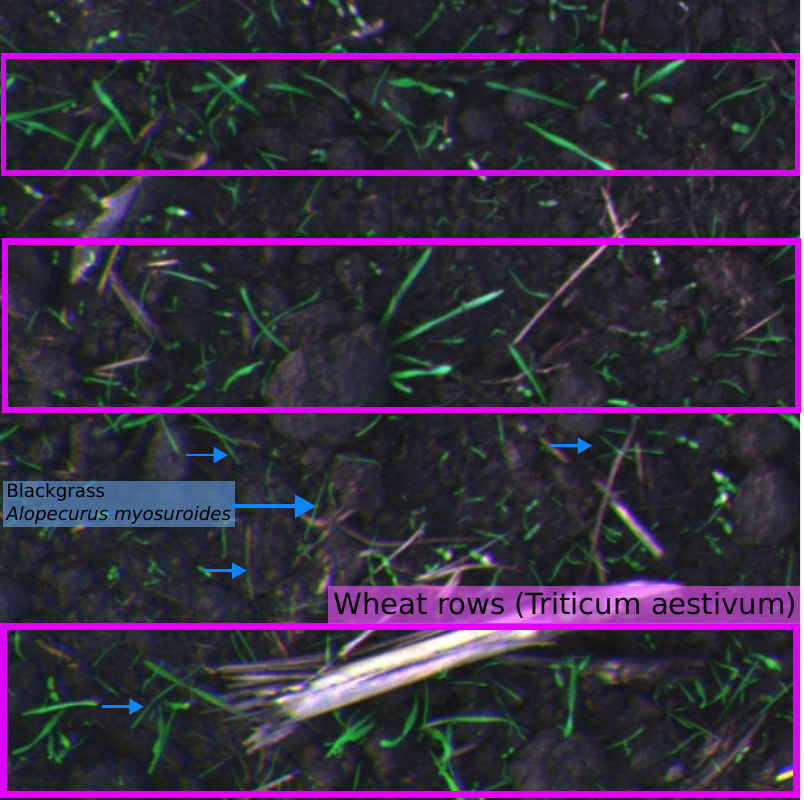}}
\subcaptionbox{No Blackgrass: Early Season \label{fig:dataset:nbg:early}}
{\includegraphics[width=0.4\linewidth]{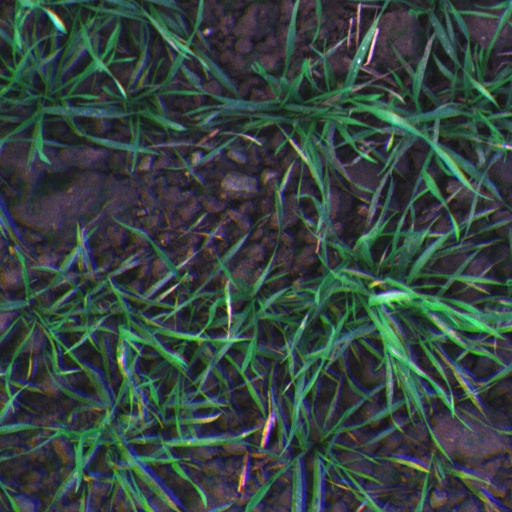}}
\subcaptionbox{Blackgrass: Late Season \label{fig:dataset:bg:late}}
{\includegraphics[width=0.4\linewidth]{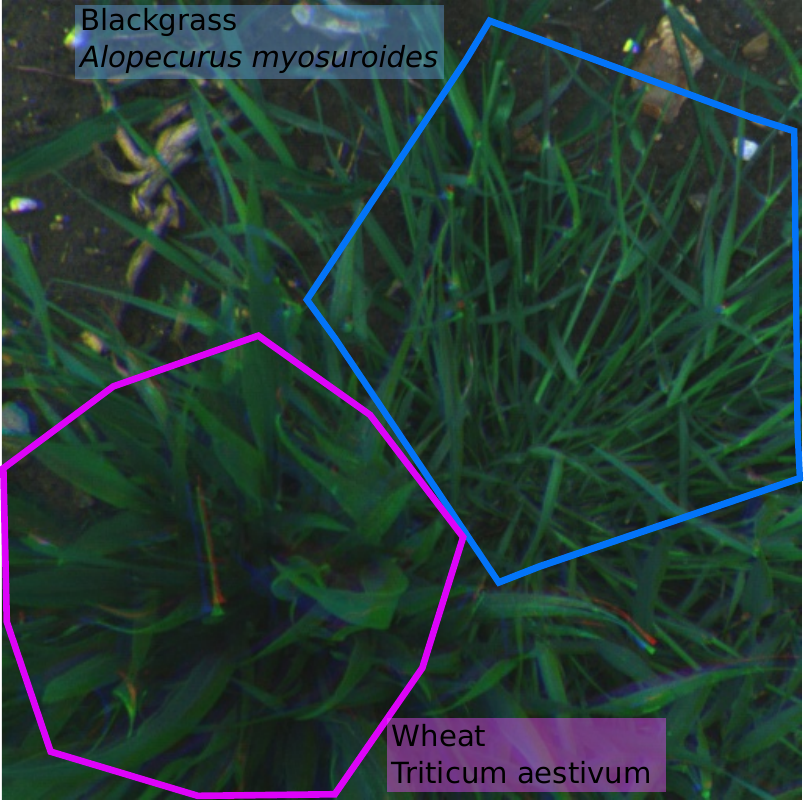}}
\subcaptionbox{No Blackgrass: Late Season \label{fig:dataset:nbg:late}}
{\includegraphics[width=0.4\linewidth]{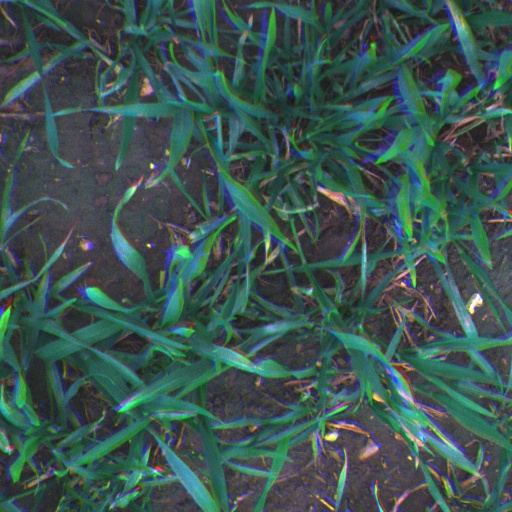}}
\caption{RGB examples from the dataset. In a) blackgrass are the thin single green strands and the wheat is the wider leaf blades. Blue arrows show a few indicative examples of blackgrass, but there are many blackgrass seedlings in a). In c), the blackgrass it the plant towards the top right, with thinner blades, while the wheat plant is towards the lower left with wider blades. b) and c) show examples without blackgrass.}
\label{fig:dataset}
\end{figure*}

Figures \ref{fig:dataset:bg:early} and \ref{fig:dataset:nbg:early} are example images from the early growing season (November to January) where the crop plants were well separated and blackgrass was present in short, singular, thin blades. Figures \ref{fig:dataset:bg:late} and \ref{fig:dataset:nbg:late} are from February on-wards where the crop canopy exposes less bare soil and the individual blackgrass blades developed into a bushier structure.

For each field, the soil type in that field was categorised according to the Cranfield Soil and Agrifood Institute's Soilscapes databases \cite{soilscapes}. 
Due to the geography of the farms that were visited, there were no fields visited that contained predominantly peat, or predominantly chalky soil. Table \ref{tab:dataset-soils} describes each soil type in the dataset and shows the break down of images with each soil type in each of the dataset splits. Note that, due to prioritising a stratification based on geo-location, it was not possible to have every soil type represented in each dataset split.

\begin{table}[t]
    \centering
    \begin{tabular}{ |p{1.5cm}|p{5cm}|p{1cm}|p{1.5cm}|p{1cm}|p{1cm}| } 
        \hline
        Soil Type & Description & Train & Validation & Test & Total \\
        \hline
        Type 5 & Free draining lime-rich loamy                   & 379  & 0    & 0    & 379  \\ \hline
        Type 7 & Freely draining slightly acid but base-rich soils & 497  & 0    & 581  & 1078 \\ \hline
        Type 8 & Slightly acid loamy and clayey soils with impeded drainage           & 489  & 0    & 492  & 981  \\ \hline
        Type 9 & Lime-rich loamy and clayey soils with impeded drainage                  & 1724 & 0    & 0    & 1724 \\ \hline
        Type 17 & Slowly permeable seasonally wet acid loamy and clayey soils                          & 6204 & 1101 & 197  & 7502 \\ \hline
        Type 18 & Slowly permeable seasonally wet slightly acid but base-rich loamy and clayey soils & 1444 & 214  & 0    & 1658 \\ \hline
        Type 23 & Loamy and sandy soils with naturally high groundwater and a peaty surface  & 1015 & 744  & 848  & 2607 \\ \hline
    \end{tabular}
    \caption{Soil types}
    \label{tab:dataset-soils}
\end{table}

\subsection{Neural Network Architectures}
For this study, we choose ResNet-50, EfficientNet B4 and Swin-B as the representative architectures for the benchmarks. They are detailed in Table \ref{tab:models}. ResNet \cite{he2016deep} introduced residual learning by incorporating skip connections, enabling the training of very deep neural networks. Despite being the oldest of the architectures, it is still widely used, due to its training efficiency and accuracy. EfficientNet \cite{tan2019efficientnet} improved image classification on ImageNet by optimizing for network size and computational efficiency using a neural architecture search approach. This was selected as another popular choice of CNN architecture but more explicitly designed for computational efficiency than ResNet. Given, the recent interest in transformers for machine vision, we chose the Swin Transformer \cite{liu2021swin} which improved on Vision Transformer (ViT) by introducing a hierarchical architecture that processes images at multiple scales in an efficient manner. This hierarchical design facilitated capturing both local and global features effectively, addressing the limitations of the single-level processing in ViT.

\begin{table}[t]
    \centering
    \begin{tabular}{ |c|c|c|c|c| } 
        \hline
        Model & Params & GFLOPS & ImageNet Acc@1 & ImageNet Acc@5 \\ \hline
        Resnet 50     & 25.6M & 4.09 & 76.13 & 92.86 \\ \hline
        Efficient Net B4 & 19.3M & 4.39 & 83.38 & 96.59  \\ \hline
        Swin B        & 87.8M & 15.43 & 83.58 & 96.64 \\ \hline
    \end{tabular}
    \caption{Models}
    \label{tab:models}
\end{table}

\subsection{Experiments}
Using these models, we perform the following experiments on our dataset:
\begin{itemize}
    \item \textbf{Baseline Models} We establish baseline results for our dataset on our selected models. Furthermore, we examine the performance on each of the separate fields in the test to understand the strengths and weaknesses of the models in different environments.
    \item \textbf{Spectral Band Importance} To gain insight into the role of each spectral band, the models were trained on all the combinations of spectral bands in the dataset.
    \item \textbf{Training Data Quantity Experiment} To assess the significance of the training data quantity for each tested model, we evaluated them on subsets containing 1\%, 3\%, 6\%, 12\%, 25\%, 50\%, and 75\% of the training data. These subsets were generated through random sampling from all training images across fields; maintaining the original proportion of positive and negative samples from the entire dataset.
\end{itemize}

\subsection{Training Settings}
The models were trained on an Nvidia GeForce RTX 2020 Ti. The initial weights for each model were taken from Pytorch's version of the models pretrained on ImageNet. Since these models were trained on RGB input, the NIR and red edge channels were set to the pretrained weights for the red channel from these models. Each model was trained for 50 epochs. The model with the highest accuracy on the validation set is selected. The optimizer was stochastic gradient descent (SGD) \cite{robbins1951stochastic}. The initial learning rate was 0.001 and this was decayed to 0.0001 using cosine annealing. The batch size used was 8. Cross entropy loss, also known as log loss, was used as the objective function to train the models. The training settings outlined were determined empirically.

\subsection{Evaluation}
In order to give a comprehensive overview of evaluation of a classification model's performance, we use four standard classification metrics: accuracy, precision, recall and MCC. Since the dataset is mostly balanced, accuracy gives a general overview of the quality of predictions. Accuracy represents that total fraction of correct predictions made by the model and is calculated as follows:
\begin{equation}
    Accuracy = \frac{TP + TN}{TP + TN + FP + FN}
\end{equation}
where $TP$ is the total number of true positives, $TN$ is the total true negatives, $FP$ is the total false positives and $FN$ is the total false negatives.

Additionally, precision and recall with respect to the blackgrass class offers insights into the model's ability to correctly identify positive instances and capture all relevant positive instances, respectively. Precision is the fraction of positive predictions that are correctly predicted and is calculated as follows:
\begin{equation}
    Precision = \frac{TP}{TP + FP}
\end{equation}
Recall is the fraction of the positive class present in the dataset correctly predicted by the model and is calculated as follows:

\begin{equation}
    Recall = \frac{TP}{TP + FN}
\end{equation}
Matthews Correlation Coefficient (MCC) is another way to assess if the quality of binary classification predictions is robust to class inbalances in the dataset. While our dataset is mostly balanced, this metric allows us to account for the slight difference in class balance. Additionally, it will enable a comparison with work published on unbalanced datasets. MCC produces values between -1 and 1 where 1 indicates perfect prediction, 0 indicates a random prediction and -1 means a perfectly inverse prediction. MCC is calculated as follows:
\begin{equation}
    MCC = \frac{TP * TN - FP * FN}{\sqrt{(TP + FP)(TP + FN)(TN + FP)(TN + FN)}}
\end{equation} 

For each of these metrics, we report the mean and 2 standard deviations of 5 training runs each with different random seeds.

\section{Results}
\begin{figure*}[t]
  \centering
  \includegraphics[width=0.8\linewidth]{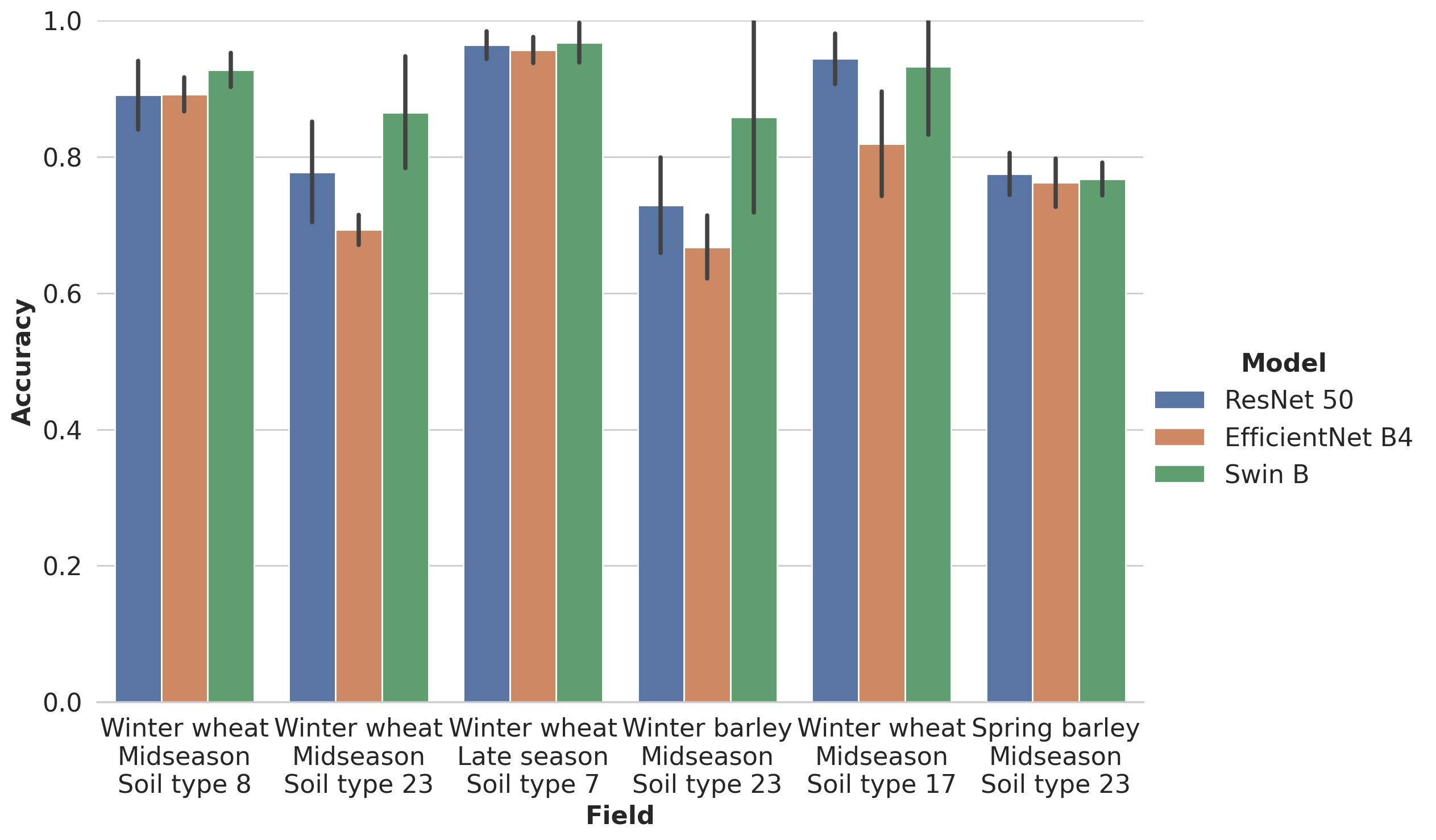}
  \caption{The accuracy of each model—ResNet50 (blue), EfficientNet B4 (orange), and Swin B (green)—on each field in the test set. The mean of 5 models from repeated training runs was taken. The error bars represent a range of two standard deviations around the mean.}
  \label{fig:field}
\end{figure*}

\begin{figure*}[t]
  \centering
  \includegraphics[width=0.7\linewidth]{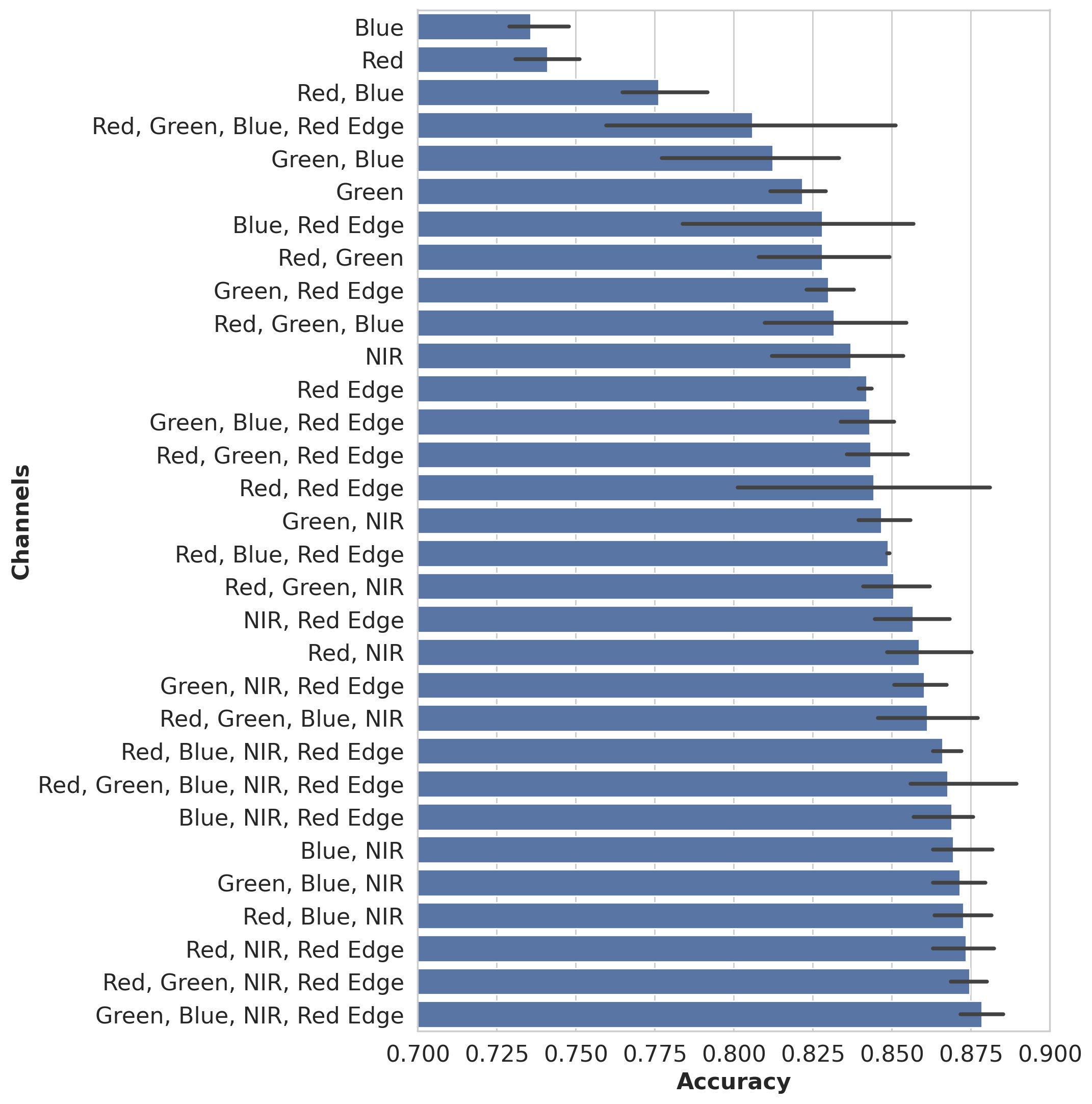}
  \caption{Models trained on different combinations of color channels. Models, identified by the combination of channels used, are ordered by accuracy. The mean of 3 models from repeated training runs was taken. The error bars represent a range of two standard deviations around the mean.}
  \label{fig:channels}
\end{figure*}

\begin{figure*}[]
  \centering
  \includegraphics[width=0.8\linewidth]{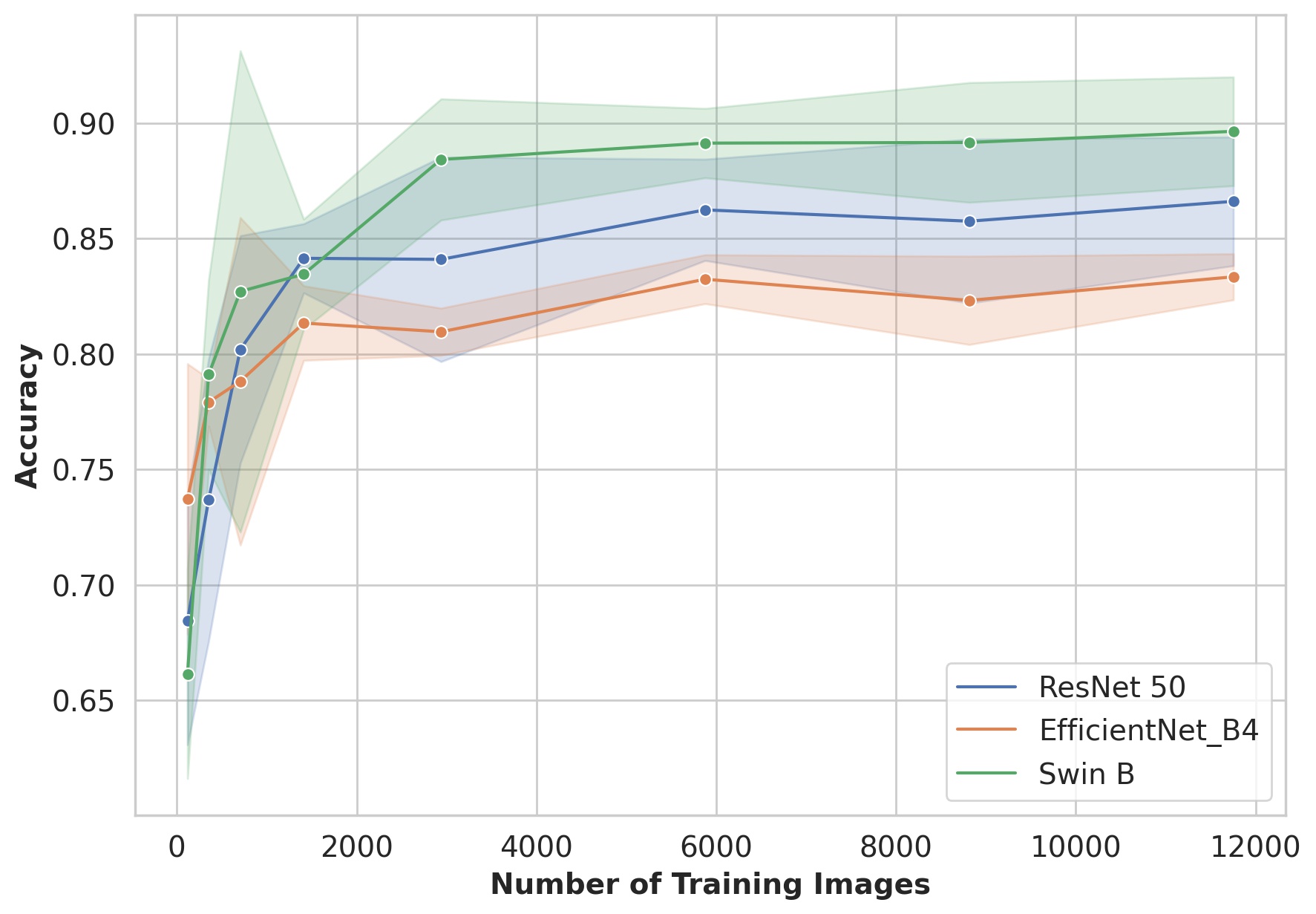}
  \caption{The accuracy of each model—ResNet50 (blue), EfficientNet B4 (orange), and Swin B (green)—with different quantities of training data. The mean of 5 models from repeated training runs was taken. The coloured bands
  represent a range of two standard deviations around the mean.}
  \label{fig:models-accuracy}
\end{figure*}

Table \ref{tab:results} shows the accuracy, MCC, precision and recall. Figure \ref{fig:field} shows the accuracy for each model on each of the different fields in the test sets. Figure \ref{fig:channels} shows the accuracy of ResNet-50 trained on every combination of spectral band. Figure \ref{fig:models-accuracy} shows the how the accuracy changes as the number of training images increases.


\begin{table}[t]
    \centering
    \begin{tabular}{|c|c|c|c|c|}
      \hline Model & Accuracy & MCC & Precision & Recall \\
      \hline
      ResNet 50 & $0.866 \pm 0.026$ & $0.855 \pm 0.010$ & $0.851 \pm 0.044$  & $0.733 \pm 0.050$ \\
      \hline
      Efficient B4 & $0.833 \pm 0.010$  & $0.869 \pm 0.036$ & $0.799 \pm 0.050$ & $0.670 \pm 0.019$\\
      \hline
      Swin B & $0.896 \pm 0.024$ & $0.912 \pm 0.051$  & $0.886 \pm 0.028$ & $0.794 \pm 0.048$ \\
      \hline
    \end{tabular}
    \caption{Baseline results. The mean of 5 models from repeated training runs was taken. Each metric is given with 2 standard deviations.}
    \label{tab:results}
\end{table}

Table \ref{tab:results} shows that all models achieved an accuracy greater than 0.8. Swin B has the highest accuracy with 0.896, however, ResNet 50 performs only slightly worse with 0.866. EfficientNet B4 performs less well than the other two models with only 0.833.

Figure \ref{fig:field} shows that the performance of the models varied considerably between the different fields. Mostly, models performed worse on barley compared with wheat and worse on midseason crops compared to late season crops.

Figure \ref{fig:channels} highlights the importance of NIR in the discrimination of wheat and blackgrass.

Figure \ref{fig:models-accuracy} shows that increasing the number training images improves performance up to the first 6000 images. From there, performance does not improve. Swin-B shows more variability in performance with a small number of images than the other models.

\section{Discussion}
Our analysis of the existing literature showed the need for more research into the challenging visual recognition task of recognising grass weeds in grass crops. The dataset we provide aims to offer an opportunity for further research in this area. Given the scale of the dataset, we hope it offers an opportunity to learn generalised solutions. In order to test a model's ability to generalise, the validation and test set are made up of images from unseen fields --- that is images from fields not in the training dataset. 

Overall, the results in Table 5 show that all models achieve an average accuracy of over 80\% using all the training data across all the test fields. We can conclude Swin B is the best performing on average. ResNet 50 is only slightly worse on average while EfficientNet B4 performs worse overall. All models exhibit high precision, indicating few false positives, but lower recall suggests more false negatives–that is, more blackgrass instances are being missed.

\subsection{Fields}
In general, Figure \ref{fig:field} shows all models performed worse on barley crops compared with wheat crops. This could be the result of fewer images of barley crops being in the dataset. Additionally, all models performed best on the fields where images were taken during the late season. Despite fewer images in the training data of late season crops, these were more accurately classified. This suggests that as the crop and blackgrass mature they become easier to discriminate.

\subsection{Channels}

The dataset is comprised of multispectral images, and allowed us to test the role different spectral bands could play in discriminating between cereal crops and blackgrass. In line with other research, we find that the addition of NIR reliably improves classification accuracy \cite{sa2018weednet, wang2020semantic}, shown in Figure \ref{fig:channels}.  For example, Red on its own has an accuracy of 0.741 while Red and NIR has an accuracy of 0.859. Similarly, Red and Blue together have an accuracy of 0.776 but with NIR the accuracy is 0.873. Furthermore, the 14 most accurate models all have NIR as an input.

Some powerful subsets like Red, Red Edge and NIR outperform the use of all channels on average, suggesting that particular discriminating features are present in a subset of spectral bands and that in some cases the addition of other spectral bands adds noise to the classification process.

\subsection{Training Data Size}
Finally, since our dataset is unusually large for an agricultural classification dataset we wanted to determine whether this larger dataset conferred any benefits. We found that up to 6000 images there was an improvement in performance but beyond that the performance improvement was small, as shown in \ref{fig:models-accuracy}. It could be that the variation in the 37 fields in the training set was well represented by the 6000 randomly sampled images and additional images provided minimal additional discriminating information. While EfficientNet has the lowest accuracy of the approaches tried, it performs the best on a very restricted dataset size.

\subsection{Limitations}
While many popular crop and weed datasets have bounding box labels or pixel-level labels, our dataset only has image-level labels. This is due to the fact that to the classification of blackgrass could often not be made by looking at the image, especially during the early season, even by an expert observer. Instead, a plant’s anatomy would need to be carefully examined in-person to make such a classification. However, we acknowledge that a weed management system trained on these images may have limited precision, as it can only recognise blackgrass at the image level.

At present, this work focuses on recognising the presence of black- grass, however, later work will investigate blackgrass density classification.

\section{Conclusion}
The problem of grass weed recognition in grass crops, including in major cereal crops, is under-represented in the research to date. To address this, we provide a large dataset of labelled images of blackgrass in wheat and barley crops and show the performance of a representative set of deep classification models in distinguishing between images with and without blackgrass. We demonstrate the performance of state-of- the-art machine vision techniques on unseen fields. We explore the usefulness of additional spectral bands in the classification as well as the importance of additional training data for different popular training models. We find NIR is important for blackgrass classification which concurs with other work that has shown it is useful for weed recognition tasks. Additionally, we establish that for our dataset and this task, there is an upper limit on the number of images that yield improvements in accuracy.

\section*{Data Availability Statement}
The dataset used in this study will be made available for non-commercial use under the Creative Commons Attribution-NonCommercial-ShareAlike (CC BY-NC-SA) license. Access to the dataset will be granted upon request, subject to agreement with the terms of the license.

\section*{Acknowledgement} 
This research was partially supported by a PhD studentship from the University of Lincoln, and by Lincoln Agri-Robotics as part of the Expanding Excellence in England (E3) Programme. We would like to acknowledge support from the UKRI Research England Ceres Agri-tech grant for funding the collection of the data set and initial testing in project Ceres:A13722 IC8P1; Autonomous Black-Grass Detection. We would also like to thanks the land managers that helped us identify fields and black-grass patches, and gave us access for image collection.

\section*{Author Contributions}
\textbf{Madeleine Darbyshire}: Conceptualization, Investigation, Methodology, Software, Visualization, Writing – original draft. \textbf{Shaun Coutts}: Conceptualization, Funding acquisition, Investigation, Supervision, Writing – review \& editing. \textbf{Eleanor Hammond}: Investigation. \textbf{Fazilet Gokbudak}: Investigation. \textbf{Cengiz Ozterli}: Funding acquisition, Supervision. \textbf{Petra Bosilj}: Conceptualization, Funding acquisition, Methodology, Writing – review \& editing. \textbf{Junfeng Gao}: Investigation. \textbf{Elizabeth Sklar}: Supervision, Writing – review \& editing. \textbf{Simon Parsons}: Supervision, Writing – review \& editing.

\appendix

\section{Full List of Wheat and Barley Papers by Weed Species}
\label{app:weed}

\begin{longtable}{ | p{0.3\linewidth} | p{0.28\linewidth} | p{0.05\linewidth} | p{0.28\linewidth} | } 
    \hline Latin Name & Common Name & Total & Publications \\ \hline
           Aethusa cynapium        & & 1 & \cite{anderegg2023farm} \\ \hline
           Alchemilla arvensis     & Parsley Piert   & 2 & \cite{anderegg2023farm, xu2023weedsnet} \\ \hline
           Alopecurus aequalis     & Shortawn Foxtail & 1 & \cite{xu2022multi} \\ \hline
           Alopecurus myosuroides  & Blackgrass  & 32 & \cite{dyrmann2015roboweedsupport, sukumar2016weed, chavan2018agroavnet, binguitcha2019crops, elnemr2019convolutional, jiang2019novel, espejo2020improving, fu2020field, rahman2020performance, skacev2020development, trong2020late, espejo2021combining, espejo2021testing, farkhani2021weed, jensen2021predictive, mounashree2021automated, siddiqui2021neural, trong2021yielding, damilare2022weed, fraccaro2022deep, kumar2022weed, kundur2022deep, mu2022faster, reddy2022weed, srivastava2022role, tara2022machine, bharathi2023plant, dheeraj2023using, guo2023identification, mishra2023comparative, mu2023densenet, rahman2023new} \\ \hline
           Amaranthus retroflexus  & Redroot Pigweed  & 4 & \cite{xu2020recognition, ronay2022effect, xu2022multi, xu2023weedsnet} \\ \hline
           Apera spica-venti       & Loose Silky-bent & 30 & \cite{dyrmann2015roboweedsupport, chavan2018agroavnet, binguitcha2019crops, elnemr2019convolutional, jiang2019novel, espejo2020improving, fu2020field, rahman2020performance, skacev2020development, trong2020late, espejo2021combining, espejo2021testing, farkhani2021weed, mounashree2021automated, siddiqui2021neural, trong2021yielding, damilare2022weed, kumar2022weed, kundur2022deep, mu2022faster, reddy2022weed, srivastava2022role, tara2022machine, anderegg2023farm, bharathi2023plant, dheeraj2023using, guo2023identification, mishra2023comparative, mu2023densenet, rahman2023new} \\ \hline
           Arctotheca calendula    & Cape Dandelion & 1 & \cite{le2021detecting} \\ \hline
           Avena fatua             & Wild Oat         & 1 & \cite{shahbazi2021assessing} \\ \hline
           Bromus japonicus        & Japanese brome & 2 & \cite{xu2022multi, xu2023weedsnet} \\ \hline
           Capsella bursa-pastoris & Shepherd’s Purse & 36 & \cite{dyrmann2015roboweedsupport, dyrmann2017roboweedsupport, chavan2018agroavnet, dyrmann2018using, binguitcha2019crops, elnemr2019convolutional, jiang2019novel, espejo2020improving, fu2020field, rahman2020performance, skacev2020development, trong2020late, xu2020recognition, espejo2021combining, espejo2021testing, farkhani2021weed, mounashree2021automated, siddiqui2021neural, trong2021yielding, damilare2022weed, kumar2022weed, kundur2022deep, mu2022faster, reddy2022weed, srivastava2022role, tara2022machine, xu2022multi, zhuang2022evaluation, anderegg2023farm, bharathi2023plant, dheeraj2023using, guo2023identification, liu2023semi, mishra2023comparative, mu2023densenet, rahman2023new} \\ \hline
           Centaurea cyanus        & Cornflower & 1 & \cite{anderegg2023farm} \\ \hline
           Chenopodium album       & Fat Hen & 35 & \cite{dyrmann2015roboweedsupport, dyrmann2017roboweedsupport, chavan2018agroavnet, dyrmann2018using, wang2018low, binguitcha2019crops, elnemr2019convolutional, jiang2019novel, espejo2020improving, fu2020field, rahman2020performance, skacev2020development, trong2020late, wang2020transfer, espejo2021combining, espejo2021testing, farkhani2021weed, mounashree2021automated, siddiqui2021neural, trong2021yielding, damilare2022weed, kumar2022weed, kundur2022deep, mu2022faster, reddy2022weed, srivastava2022role, tara2022machine, alirezazadeh2023comparative, anderegg2023farm, bharathi2023plant, dheeraj2023using, guo2023identification, mishra2023comparative, mu2023densenet, rahman2023new} \\ \hline
           Cerastium fontanum & Common Mouse Ear Chickweed & 1 & \cite{anderegg2023farm} \\ \hline
           Chloris cucullata       & Hooded Windmill Grass & 1 & \cite{saqib2023towards} \\ \hline
           Cirsium arvense         & Creeping Thistle & 5 & \cite{franco2018automatic, rasmussen2019pre, rasmussen2021pre, haq2023weed, saqib2023towards, thomas2023weakly} \\ \hline
           Consolida regalis       & Forking Larkspur & 3 & \cite{almalky2022efficient, almalky2023deep, almalky2023real} \\ \hline
           Convolvulus arvensis    & Field Bindweed & 3 & \cite{raptis2022multimodal, anderegg2023farm, saqib2023towards} \\ \hline
           Dactylis glomerata      & Cat Grass & 1 & \cite{saqib2023towards} \\ \hline
           Echinochloa crus-galli  & Cockspur Grass & 2 & \cite{xu2022multi, xu2023weedsnet} \\ \hline
           Elymus repens           & Common Couch & 1 & \cite{thomas2023weakly} \\ \hline
           Equisetum arvense       & Horsetail & 1 & \cite{zou2021modified} \\ \hline
           Eschscholzia californica & California poppy & 1 & \cite{saqib2023towards} \\ \hline
           Fallopia convolvulus & Black Bindweed & 1 & \cite{anderegg2023farm} \\ \hline
           Fumaria officinalis & Common fumitory & 1 & \cite{anderegg2023farm} \\ \hline
           Galium aparine          & Cleavers    & 34 & \cite{dyrmann2015roboweedsupport, dyrmann2017roboweedsupport, chavan2018agroavnet, dyrmann2018using, binguitcha2019crops, elnemr2019convolutional, jiang2019novel, espejo2020improving, fu2020field, rahman2020performance, skacev2020development, trong2020late, espejo2021combining, espejo2021testing, farkhani2021weed, mounashree2021automated, siddiqui2021neural, trong2021yielding, damilare2022weed, kumar2022weed, kundur2022deep, mu2022faster, reddy2022weed, srivastava2022role, tara2022machine, zhuang2022evaluation, anderegg2023farm, bharathi2023plant, dheeraj2023using, guo2023identification, liu2023semi, mishra2023comparative, mu2023densenet, rahman2023new} \\ \hline
           Galium odoratum         & Sweet Woodruff & 1 & \cite{liu2023semi} \\ \hline
           Geranium pusillum       & Small-flowered Cranesbill & 30 & \cite{dyrmann2017roboweedsupport, chavan2018agroavnet, dyrmann2018using, binguitcha2019crops, elnemr2019convolutional, jiang2019novel, espejo2020improving, fu2020field, rahman2020performance, skacev2020development, trong2020late, espejo2021combining, espejo2021testing, farkhani2021weed, mounashree2021automated, siddiqui2021neural, trong2021yielding, damilare2022weed, kumar2022weed, kundur2022deep, mu2022faster, reddy2022weed, srivastava2022role, tara2022machine, bharathi2023plant, dheeraj2023using, guo2023identification, mishra2023comparative, mu2023densenet, rahman2023new} \\ \hline
           Humulus japonicus       & Japanese Hops & 2 & \cite{wang2018low, wang2020transfer} \\ \hline
           Lamium purpureum        & Red Deadnettle & 2 & \cite{alirezazadeh2023comparative, anderegg2023farm} \\ \hline
           Lolium multiflorum      & Italian Ryegrass & 2 &  \cite{jensen2021predictive, anderegg2023farm} \\ \hline
           Lolium perenne          & Perennial Ryegrass & 3 & \cite{dyrmann2017roboweedsupport, dyrmann2018using, saqib2023towards} \\ \hline
           Lipandra polysperma     & Manyseed Goosefoot & 1 & \cite{anderegg2023farm} \\ \hline
           Lithospermum arvense    & Corn Gromwell & 1 & \cite{alirezazadeh2023comparative} \\ \hline
           Matricaria chamomilla   & Chamomile & 1 & \cite{anderegg2023farm} \\ \hline
           Matricaria discoidea    & Pineappleweed & 2 & \cite{sukumar2016weed, pflanz2018weed} \\ \hline
           Myosotis arvensis       & Field Forget-Me-Not & 3 & \cite{dyrmann2017roboweedsupport, dyrmann2018using, anderegg2023farm} \\ \hline
           Papaver rhoeas          & Common Poppy & 3 & \cite{pflanz2018weed, anderegg2023farm, perez2023early} \\ \hline
           Persicaria maculosa     & Redshank & 3 & \cite{dyrmann2017roboweedsupport, dyrmann2018using, anderegg2023farm} \\ \hline
           Phalaris paradoxa       & Awned Canarygrass & 1 & \cite{jabir2021accuracy} \\ \hline
           Poa annua               & Annual Bluegrass & 3 & \cite{xu2020recognition, xu2022multi, anderegg2023farm} \\ \hline
           Poa pratensis           & Kentucky bluegrass & 1 & \cite{xu2023weedsnet} \\ \hline
           Polygonum aviculare     & Common Knotweed & 1 & \cite{anderegg2023farm} \\ \hline
           Portulaca oleracea      & Purslane & 1 & \cite{raptis2022multimodal} \\ \hline 
           Raphanus raphanistrum   & Wild Radish & 2 & \cite{le2021detecting, shahbazi2021assessing} \\ \hline
           Rumex crispus           & Curly Dock & 2 & \cite{el2022metaheuristic, pp2023metaheuristic} \\ \hline
           Rumex obtusifolius      & Broadleaf Dock & 2 & \cite{el2022metaheuristic, pp2023metaheuristic} \\ \hline
           Setaria adhaerens       & & 1 & \cite{ronay2022effect} \\ \hline
           Setaria viridis         & Green Foxtail & 1 & \cite{zou2021modified} \\ \hline
           Sinapis arvensis        & Charlock    & 31 & \cite{dyrmann2015roboweedsupport, dyrmann2017roboweedsupport, chavan2018agroavnet, dyrmann2018using, binguitcha2019crops, elnemr2019convolutional, jiang2019novel, espejo2020improving, fu2020field, rahman2020performance, skacev2020development, trong2020late, espejo2021combining, espejo2021testing, farkhani2021weed, mounashree2021automated, siddiqui2021neural, trong2021yielding, damilare2022weed, kumar2022weed, kundur2022deep, mu2022faster, reddy2022weed, srivastava2022role, tara2022machine, bharathi2023plant, dheeraj2023using, guo2023identification, mishra2023comparative, mu2023densenet, rahman2023new} \\ \hline
           Sorghum halepense       & Johnson grass & 1 & \cite{raptis2022multimodal} \\ \hline
           Solanum nigrum          & Black Nightshade & 4 & \cite{dyrmann2017roboweedsupport, dyrmann2018using, ronay2022effect, anderegg2023farm} \\ \hline
           Spergula arvensis       & Corn Spurry & 1 & \cite{alirezazadeh2023comparative} \\ \hline
           Stellaria media         & Common Chickweed & 35 & \cite{dyrmann2015roboweedsupport, dyrmann2017roboweedsupport, chavan2018agroavnet, dyrmann2018using, binguitcha2019crops, elnemr2019convolutional, jiang2019novel, espejo2020improving, fu2020field, rahman2020performance, skacev2020development, trong2020late, espejo2021combining, espejo2021testing, farkhani2021weed, mounashree2021automated, siddiqui2021neural, trong2021yielding, damilare2022weed, kumar2022weed, kundur2022deep, mu2022faster, reddy2022weed, srivastava2022role, tara2022machine, zhuang2022evaluation, alirezazadeh2023comparative, anderegg2023farm, bharathi2023plant, dheeraj2023using, guo2023identification, liu2023semi, mishra2023comparative, mu2023densenet, rahman2023new} \\ \hline
           Taraxacum officinale    & Common Dandelion & 1 & \cite{anderegg2023farm} \\ \hline
           Trifolium pratense      & Red Clover & 2 & \cite{dyrmann2017roboweedsupport, dyrmann2018using} \\ \hline
           Tripleurospermum inodorum & Scentless Mayweed & 32 & \cite{dyrmann2015roboweedsupport, dyrmann2017roboweedsupport, chavan2018agroavnet, dyrmann2018using, binguitcha2019crops, elnemr2019convolutional, jiang2019novel, espejo2020improving, fu2020field, rahman2020performance, skacev2020development, trong2020late, espejo2021combining, espejo2021testing, farkhani2021weed, mounashree2021automated, siddiqui2021neural, trong2021yielding, damilare2022weed, kumar2022weed, kundur2022deep, mu2022faster, reddy2022weed, srivastava2022role, tara2022machine, bharathi2023plant, dheeraj2023using, guo2023identification, mishra2023comparative, mu2023densenet, rahman2023new, thomas2023weakly} \\ \hline
           Urtica urens            & Small Nettle & 2 & \cite{dyrmann2017roboweedsupport, dyrmann2018using} \\ \hline
           Veronica hederifolia    & Ivy-leaved Speedwell & 2 & \cite{pflanz2018weed, anderegg2023farm} \\ \hline
           Veronica persica        & Common field-speedwell & 5 & \cite{sukumar2016weed, dyrmann2017roboweedsupport, dyrmann2018using, anderegg2023farm, liu2023semi} \\ \hline
           Viola arvensis          & Field Pansy & 4 & \cite{dyrmann2017roboweedsupport, dyrmann2018using, pflanz2018weed, anderegg2023farm} \\ \hline
           Vulpia myuros           & Rattail Fescue & 1 & \cite{jensen2021predictive} \\ \hline
           Xanthium strumarium     & & 2 & \cite{wang2018low, wang2020transfer} \\ \hline
\end{longtable}

\bibliographystyle{plain}
\bibliography{references}

\end{document}